# Structured Region Graphs: Morphing EP into GBP

**Max Welling**
Dept. of Computer Science
UC Irvine
Irvine CA 92697-3425
*welling@ics.uci.edu*

**Thomas P. Minka**
Microsoft Research
Cambridge, CB3 0FB, UK
*minka@microsoft.com*

**Yee Whye Teh**
Computer Science Division
UC Berkeley
Berkeley CA 94720-1776
*ywteh@eecs.berkeley.edu*

## Abstract

GBP and EP are two successful algorithms for approximate probabilistic inference, which are based on different approximation strategies. An open problem in both algorithms has been how to choose an appropriate approximation structure. We introduce "structured region graphs," a formalism which marries these two strategies, reveals a deep connection between them, and suggests how to choose good approximation structures. In this formalism, each region has an internal structure which defines an exponential family, whose sufficient statistics must be matched by the parent region. Reduction operators on these structures allow conversion between EP and GBP free energies. Thus it is revealed that all EP approximations on discrete variables are special cases of GBP, and conversely that some well-known GBP approximations, such as overlapping squares, are special cases of EP. Furthermore, region graphs derived from EP have a number of good structural properties, including maxent-normality and overall counting number of one. The result is a convenient framework for producing high-quality approximations with a user-adjustable level of complexity.

## 1 INTRODUCTION

One of the most successful algorithms for approximate inference is the generalized belief propagation (GBP) algorithm and its corresponding "region-based" approximation Yedidia et al. (2002) (see also McEliece & Yildirim (1998) and Pakzad & Anantharam (2003)). Independently, the expectation propagation (EP) algorithm was introduced in Minka (2001a) as an approximate Bayesian inference method. When restricted to a fully factorized structure, EP was shown to be equivalent to loopy belief propagation. The case of tree structures was discussed by Minka & Qi (2004) and found to have some parallels with GBP.

In the design of EP algorithms the emphasis has been more on algorithmic considerations (though Minka (2001a; 2001b) did define an EP free energy function). On the other hand, the design of GBP algorithms seems to be mostly guided by constructing good approximations to the free energy of the problem. These two approaches have perhaps contributed to the fact that no clear relationship between them has previously been uncovered.

Exposing this relationship is important for the following reasons. The region graph framework of GBP is extremely flexible which comes at a certain cost: not every region-based approximation is accurate and it is not well understood what regions to choose in order to achieve high fidelity approximations. We show that the EP framework can usefully guide the choice of region graph. Conversely, insights from the region graph framework can provide important information as to which approximation is being made by an EP algorithm.

The "structured region graph" (SRG) formalism that we propose is an extension of the region graph formalism in Yedidia et al. (2002) to structured message passing algorithms such as EP. Both EP and GBP approximations can be accommodated in this framework while a collection of "graphical reduction operators" can morph EP-graphs into equivalent region graphs. Moreover, we show that the region graph properties that were identified in Yedidia et al. (2002) to correlate with accurate approximations are automatically satisfied for EP. This provides a convenient framework for designing high quality region graphs.

## 2 GBP and EP

Generalized belief propagation (GBP) and expectation propagation (EP) are approximate algorithms for inference in distributions that can be divided into simple factors:

$$p(\mathbf{x}) \propto \prod_a f_a(x_a) \qquad (1)$$

The goal of either algorithm is to estimate various statistics of $x$. In this paper we focus on estimating marginals of discrete random variables. The factors of $p$ can be represented by a factor graph. The GBP and EP approximations can be also represented by graph, a *structured region graph*, as illustrated in the following.

Consider a distribution on 8 variables whose factors correspond to the edges of a $2 \times 4$ grid. Belief propagation passes messages in the graph shown in figure 1 (bottom left). In the top layer we have the factors (depicted as edges) and the variables in their arguments while in the bottom layer we have the single variables (depicted as nodes). Each vertex in this graph will be called a *region*. Each region stores a local distribution over its variables. To send a downward message, the parent region marginalizes its distribution onto the child variable, divides by the child's distribution, and sends the ratio. The child then updates its distribution and relays the message to its other parents. These messages can be viewed as enforcing consistency constraints among the top-layer distributions, namely that they must have the same single-variable marginals. When these constraints are satisfied, the algorithm stops.

An alternative interpretation is given by EP (Minka, 2001a). In this interpretation, we use a more general type of graph which allows the region distributions to include *all* variables but constrains them to be partially factorized (figure 1 (top left)). The child region is fully factorized, and the parent regions are factorized except for a single edge. Downward messages again consist of marginals, which are interpreted as a *projection* of the parent's distribution onto the child's fully factorized structure. The algorithm stops when all parents have the same single-variable marginals. The results are exactly the same as BP.

Going further, consider a distribution on 16 variables whose factors correspond to the edges of a $4 \times 4$ grid, to which we apply generalized belief propagation with overlapping $2 \times 2$ square clusters (Yedidia et al., 2002). This algorithm passes messages in the three-layer region graph shown in figure 1 (bottom right). Downward messages consist of marginalizing the parent's distribution onto the child's variables (this will be either a single-variable marginal in the bottom layer or a pairwise marginal in the middle layer). Upward mes-sages relay information from other parents or (in the case of the middle layer) from child regions. These messages can be interpreted as enforcing consistency constraints; the middle-layer regions must have the same single-variable marginals and the top-layer regions must have the same pairwise marginals where they overlap.

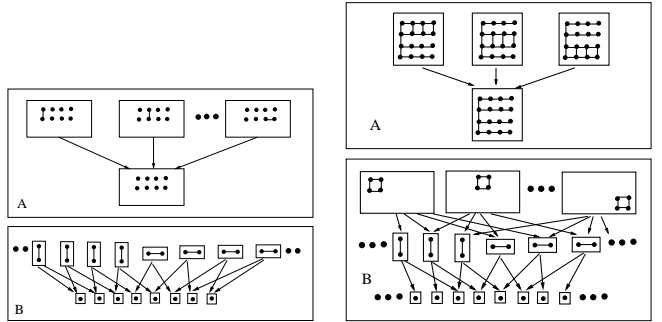

Figure 1: Left: (bottom) The region graph corresponding to belief propagation and (top) the equivalent EP-graph. Right: (bottom) The region graph described in Yedidia et al. (2002) and (top) the equivalent EP-graph.

A closely related algorithm can be obtained via tree-structured EP (Minka & Qi, 2004). It can be viewed as passing messages in a region graph, where the region distributions include all variables but are partially tree-structured (figure 1 (top right)). The child region is tree-structured, and the parent regions have the same structure plus additional edges. Downward messages consist of pairwise and single-variable marginals, which are interpreted as a projection of the parent's distribution onto the child's tree structure. The algorithm stops when all parents have the same pairwise and single-variable marginals along the tree. As we will show, the free energy of this EP algorithm is identical to the free energy of the GBP algorithm, and therefore they have the same fixed points. We will show this equivalence using purely graphical operations on the region graphs.

## 3 STRUCTURED REGION GRAPHS

This section formally defines the structured region graphs used in section 2. The common theme of GBP and EP is approximating $p(\mathbf{x})$ in a distributed fashion. Both algorithms employ a set of regions $R$, each with an associated approximate distribution $q_R(\mathbf{x})$, constrained to be in some specific family $\mathcal{H}(R)$. Each region contains some of the factors of $p$—just enough to remain tractable. The regions are tied together by constraints, e.g. they must have the same single-variable marginals. The parameters of the region distributions are optimized to meet the constraints. In setting up these algorithms, many choices have to be made: the number of regions, the allocation of factors to regions, how many parameters the regions should have, which pairs of regions should be constrained, and with what constraints. Structured region graphs are a general formalism to represent these choices.

A structured region graph (SRG) is a directed acyclic graph of regions. Regions with no parents are called *outer* regions, otherwise they are called *inner* regions. Each region $R$ has an associated set of variables $x_R$, a set of factors $f_R$ (outer regions only), a set of cliques $\mathcal{C}(R)$, and an exponential family $\mathcal{H}(R)$. Every variable in $x_R$ must appear in some clique or factor. Every factor of $p$ is assigned to a single outer region. Note that factors are separate from cliques: if there is a factor $f(x_1, x_2)$ there need not be a clique containing $x_1$ and $x_2$. From the factors and cliques we define a *structure* $\mathcal{G}(R)$, which is an undirected graph of variables linking any two variables appearing in the same factor or clique. Note that this graph can be ambiguous, e.g. a triangle might represent one clique or three separate edge cliques.

The exponential family $\mathcal{H}(R)$ is a set of distributions parameterized by $\lambda_j, j = 1, ..., J$, of the form

$$q(x_R) \propto f_R(x_R) \exp(\sum_j \lambda_j h_j(x_{C_j})) \qquad (2)$$

where $f_R(x_R) = \prod_{a \in R} f_a(x_a)$ are the factors in region $R$ and $C_j$ is one of the cliques in $\mathcal{C}(R)$. The functions $h_j$ are fixed and called *features* of the family. The expectation over $q$ of a feature is called a *moment*. This paper focuses on discrete variables where the exponential family simply has one binary feature for every configuration of variables in each clique. In this case, the family is completely specified by the cliques alone.

The region graphs of Yedidia et al. (2002) are the special case where the inner regions are complete (they contain one clique over all variables in the region).

Let $\text{pa}(R)$ be the parents of $R$, $\text{an}(R)$ the ancestors, and $\text{ch}(R)$ be the children. If $R$'s exponential family contains all of the features of another region $D$, either directly or by taking linear combinations, we say that $R$ *subsumes* $D$. It the discrete case, this means every clique in $D$ is contained in some clique of $R$ ($R$'s structure is a supergraph of $D$'s), and consequently $R$ contains all the variables of $D$.

A valid SRG must satisfy some conditions, analogous to those in Yedidia et al. (2002). For each region $R$, define a counting number $c_R$ by the recursion

$$c_R = 1 - \sum_{A \in \text{an}(R)} c_A \qquad (3)$$

Define $RG(i)$ to be the subRG of regions containing the variable $x_i$.

**Connectedness** $RG(i)$ must be connected for all $i$.

**Balancedness** $\sum_{R \in RG(i)} c_R = 1$ for all $i$.

**Hierarchy** Every region must subsume its children.

The connectedness property implies that if the factor graph of $p(\mathbf{x})$ is connected, then a valid SRG is also connected. For the rest of the paper we will assume $p(\mathbf{x})$ is connected.

Expectation propagation can be represented by a subclass of SRGs called *EP-graphs*. An EP-graph is a two-layer SRG with $A$ outer regions parenting a single inner region called the *base region*. Each region has the same set of cliques, the union of which covers all the variables of $p(\mathbf{x})$ (see figure 1).

## 4 FREE ENERGY

For each structured region graph, we can associate a free energy function which applies to both GBP and EP. Each region maintains a belief function $q_R(x_R)$. The belief functions must satisfy the following constraints:

$$\sum_{x_R \setminus x_C} q_R(x_R) = \sum_{x_D \setminus x_C} q_D(x_D) \qquad (4)$$

$$\forall D \in \text{ch}(R), \forall C \in \mathcal{C}(D) \qquad \sum_{x_R} q_R(x_R) = 1 \quad (5)$$

In other words, the marginals of a region's belief function must agree with those of the child on the child's cliques. Another way to say this is that parents and children must have the same expected value for the child's features. The Kikuchi free energy corresponding to a set of belief functions on a structured region graph is

$$F(q\|p) = \sum_R c_R \sum_{x_R} q_R(x_R) \log \frac{q_R(x_R)}{f_R(x_R)} \qquad (6)$$

subject to (4,5)

Introducing a Lagrange multiplier $\lambda_{RDC}(x_C)$ for each constraint (4), we obtain the fixed-point conditions

$$q_R(x_R) \propto f_R(x_R) \exp(\nu_R(x_R)) \qquad (7)$$

$$\nu_R(x_R) = \frac{1}{c_R} \sum_{D \in \text{ch}(R)} \sum_{C \in \mathcal{C}(D)} \lambda_{RDC}(x_C)$$

$$- \frac{1}{c_R} \sum_{A \in \text{pa}(R)} \sum_{C' \in \mathcal{C}(R)} \lambda_{ARC'}(x_{C'}) \qquad (8)$$

From the hierarchy condition, we know that each $C$ in the first line is a subset of some $C'$ in the second line.

Therefore, at a fixed point, $q_R$ factorizes according to $\mathcal{G}(R)$, or more generally each region belief is in the exponential family $\mathcal{H}(R)$.

Fixed points of the SRG free energy can be found by a message-passing algorithm that combines GBP and EP. In GBP, the parent marginalizes itself onto the child's variable set. In the new algorithm, the parent projects its belief onto the child's exponential family, by matching moments (this type of algorithm was previously described by Heskes & Zoeter, 2003). Because the messages must be in the exponential family of the child, they can be considerably simpler than GBP messages. For example, the messages may factorize according to a spanning tree. We will not concern ourselves with the general algorithm here; instead we will focus on those SRGs which reduce to ordinary RGs, by a formal process defined in the next section.

## 5 REDUCTION RULES

This section defines graphical operators on SRGs that can be used to prove the equivalence of different message-passing algorithms, via their free energies. Each of the operators preserves the critical points of the free energy, and therefore the fixed points of message-passing. They extend the operators of Welling (2004). Proofs appear in Welling et al. (2005).

Given a region $R$ and a set of variables $V$, the *induced subregion* $R[V]$ is a region possessing all cliques and factors from $R$ which only involve variables in $V$.

**Link-Death** Links into children that are also descendants via other paths can be deleted. Additionally, the link $R \to D$ can be deleted if (1) $R$ has a common ancestor ($A$) with one of $D$'s other parents (forming a loop) and (2) $D$'s other parents have no ancestor in common with $R$ except $A \cup \text{an}(A)$.

**Grow/Shrink** This applies to outer regions. New cliques can be added to any outer region, without changing the free energy (because only the child's cliques matter). This is useful for preparing an outer region to split. Similarly, cliques that have no corresponding child clique may be made smaller or deleted. This operation is independent of any factors in the region.

**Drop** This applies to a region with a unique parent and therefore zero counting number. Remove the region and link its parent directly to all of its children. (A region with more than one parent cannot be dropped, because it enforces a constraint.)

**Factor-Move** Consider two outer regions and their child region containing a clique $C$. A factor in one of the parents that is covered by this clique (i.e. the variables in the factor are a subset of the variables in the clique) can be moved to the other parent region. This is useful to prepare a Merge operation (which is only possible if the parent region contains no factors).

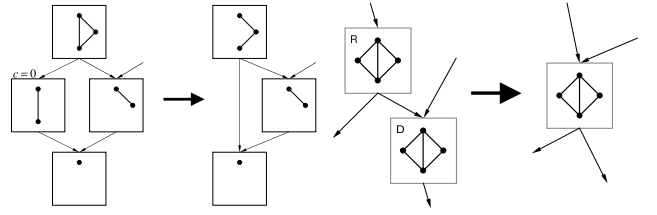

Figure 2: Drop + Shrink    Figure 3: Merge

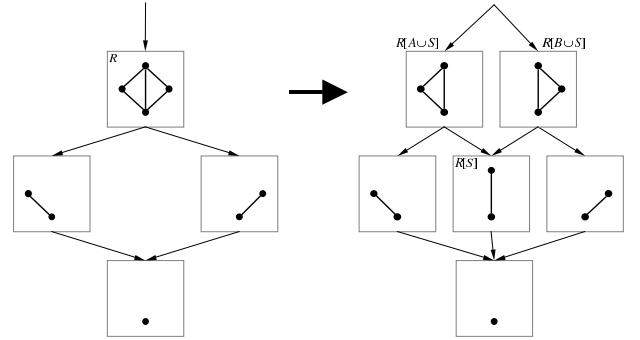

Figure 4: Split

**Merge** A parent $R$ and child $D$ can be merged under the following conditions: (1) $R$ has no factors, (2) $R$ and $D$ subsume each other, and (3) $R$'s other children (if any) have no ancestors in common with $D$ except for $R \cup \text{an}(R)$. Merge the two regions into one region which has the union of their parents, the union of their children and the sum of their counting numbers.

**Split** A split involves partitioning the variables of a region $R$ into $(A, B, S)$ such that $S$ separates $A$ from $B$ in $\mathcal{G}(R)$ ($S$ may be empty if $A$ and $B$ are disconnected). Replace $R$ with three induced subregions: $R[A \cup S]$, $R[B \cup S]$, and $R[S]$. For this to be valid, (1) $R[S]$ must be complete, and (2) for any child, all of its cliques appear together in $R[A \cup S]$ or $R[B \cup S]$. Make $R[S]$ a child of $R[A \cup S]$ and $R[B \cup S]$. Every factor in $R$ must be assigned to one of $R[A \cup S]$ or $R[B \cup S]$, with $R[S]$ stripped of all factors. Make $R[S]$ a parent of all descendants of $R$ which it subsumes, i.e. their cliques are subsets of cliques in $\mathcal{C}(R[S])$. The remaining children of $R$ that did not receive links from $R[S]$ must be subsumed by $R[A \cup S]$ or $R[B \cup S]$: make them children of $R[A \cup S]$ or $R[B \cup S]$ accordingly. The parents of $R$ become parents of both $R[A \cup S]$ and $R[B \cup S]$.

## 6 REDUCING SRGs TO RGs

Let a set of cliques be *decomposable* if they are the maximal cliques of a triangulated graph. This section shows that SRGs with decomposable inner regions reduce to ordinary RGs. As a consequence, any EP algorithm with a decomposable base region (such as a tree) is equivalent to some GBP algorithm (in the sense of having the same fixed points).

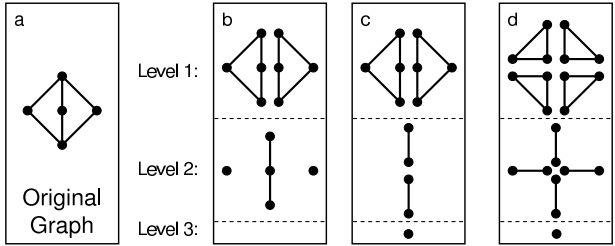

Figure 5: Reducing an EP-graph (b) to ordinary region graphs (c, d).

Figure 5 gives an example of such a reduction. Panel A is a pairwise Markov random field with one possible SRG given in panel B. The SRG is an EP-graph with two loop outer regions and a tree-structured inner region. The base region can be split into its three components and the isolated nodes dropped (they have one parent each). The middle part of the base region can be split via the middle node, giving an ordinary region graph (c). Going further, we can split the outer regions into smaller loops (d).

**Theorem 1** If an inner region $R$ is decomposable, not complete, and its children are all complete, then there exists a split of $R$ into decomposable regions.

**Proof:** If $R$'s children are complete then the second condition of a split is always fulfilled. If $R$ is decomposable and not complete then it must have a complete separator into decomposable regions (Lauritzen, 1996 sec. 2.1.2), fulfilling the first condition. □

**Theorem 2** If all inner regions are decomposable, then by reduction we can make all inner regions complete.

**Proof:** Starting at the bottom of the graph, find the first inner region which is not complete and has complete children or no children. By Th.1 this inner region can be split into decomposable regions. Eventually we are left with only complete inner regions. □

For example, consider figure 1 (left). To reduce the EP-graph, we first split the base region into singletons, then split each outer region into singletons plus one edge. Duplicate singletons merge together, leaving the conventional factor graph. In figure 1 (right), we split the tree base region into edges and nodes, then each outer region into edges, nodes, and a row of loops. Duplicate edges and nodes merge together. The loops become exactly the squares in the GBP region graph, with the same edge constraints.

Note that two algorithms which have the same fixed points do not necessarily take the same time. In practice, EP seems to be more efficient than GBP (Minka & Qi, 2004).

## 7 PROPERTIES OF SRGs

What makes a "good" SRG? Yedidia et al. (2002) described two properties of a good region graph: maxent-normality and perfect correlation resulting in $\sum_R c_R = 1$. We extend perfect correlation to SRGs and introduce a stronger condition than maxent-normality called non-singularity. Using the reduction operators, we can identify broad classes of graphs which have these properties.

### 7.1 Perfect correlation

Yedidia et al. (2002) argue that it is good for the sum of all counting numbers to equal one, because this makes the free energy exact when the variables are perfectly correlated. This argument carries over to structured region graphs, provided each region contains one connected component. In figure 1 (top right), the regions have one connected component, but in figure 1 (top left), they consist of multiple components. Thus we cannot apply their argument to the top left graph directly, but we can apply reductions until their argument does apply. The following are proved in Welling et al. (2005):

**Theorem 3** All reduction operators preserve $\sum_R c_R$ except Split with an empty separator.

**Theorem 4** An acyclic SRG (having no undirected cycles of regions) satisfies $\sum_R c_R = 1$.

Because EP-graphs are acyclic, they satisfy $\sum_R c_R = 1$. If the base approximation has one connected component, then the free energy of the EP-graph (and any reduced graph) must be exact when variables are perfectly correlated.

### 7.2 Non-singularity

The opposite of perfect correlation is when all factors are uniform. In this case, the optimal beliefs should be uniform. If the free energy is maximum at uniform beliefs, then the region graph is said to be maxent-normal (Yedidia et al., 2002). We strengthen this concept:

**Definition** A SRG is *non-singular* if message passing with uniform factors has a unique fixed point, at which all belief functions $q_R(x_R)$ are uniform.

Because all reduction operators preserve fixed points,

non-singularity is preserved by all reduction operators.

**Theorem 5** An acyclic SRG is non-singular.

**Proof:** It is possible to reduce the SRG until only single variable regions remain (which are clearly non-singular). First remove all factors. A region with no neighbors can be shrunk and split into single variable regions. Regions with neighbors can be reduced as follows. Because the graph is acyclic, at least one region has a single neighbor. If this is a child region of some parent (i.e. it is a leaf region) we have $c_R = 0$ and we can drop it. If it is parent region of some child then it is an outer region and we shrink the region to contain only the cliques of its child. We then split off all single variable regions until the structure matches the child, followed by a merge. □

Because EP-graphs are acyclic, they are non-singular and hence maxent-normal, and so is any reduction of an EP-graph. The same technique can be applied to determine whether a general (cyclic) SRG is non-singular. Remove all factors and apply the reduction operators. Ultimately, either the SRG is reduced to single variable regions, in which case the corresponding SRG is non-singular, or the SRG still has inner regions, yet cannot be further reduced. If the SRG has complete inner regions (an ordinary RG), then in the latter case the RG is singular:

**Theorem 6** If a factorless RG has complete inner regions yet is not reducible, then GBP has a fixed point other than uniform beliefs.

## 8 LOOP-GRAPHS

The previous section showed that EP-graphs automatically enjoy good properties. This section considers a more general class of SRGs and derives conditions under which they will be "good" approximations. A *loop-graph* is a SRG consisting of loop regions, edge regions, and node regions, where all structures are distinct and each loop is a parent of its constituent edges (e.g. figure 1 (bottom right), figure 5c,d). It is assumed that the factors are all pairwise, and they are assigned to loop regions or outer edge regions that are not contained in any loop. There may be more edge regions than there are factors, i.e. edge regions can simply represent overlaps. The Bethe approximation is the special case where there are no loop regions.

**Theorem 7** A loop-graph has $\sum_R c_R = L - E + V$, where $L$ is the number of loop regions, $E$ is the number of edge regions, and $V$ is the number of variables.

**Theorem 8** A loop-graph is singular iff there is a subset of loops and their constituent edges such that each of these edges is a child of at least 2 loops.

**Proof:** ⇐ Even if the other loops are removed, this subset cannot be reduced to singletons. Because all edges have two parents, they cannot be dropped. The loops cannot be shrunk nor merged. Large loops can be split into smaller loops, but their edges will still be shared. ⇒ If there is no such subset, then we can reduce to singletons as follows. Remove all factors. If there are no loop regions, then the graph reduces easily via shrink. Otherwise, there always exists a loop and an edge where the edge is unique to the loop. Drop the edge and remove its clique from the loop. The loop region becomes a tree and can be split into its nodes and edges, which subsequently merge with its descendants. Repeat on the remaining loops. □

For example, the complete graph $K_4$ has four induced loops: ABC, ABD, ACD, and BCD. A loop-graph with these loops as regions is singular. Similarly, the graph $K_{2,3}$ (depicted in figure 5a) has three induced loops which would, if all included, result in a singular loop-graph. This is related to a concept known as the *cycle space* of a graph (Diestel, 2000), analogous to the column space of a matrix. A set of cycles is said to be *linearly dependent* if their constituent edges are used an even number of times. Thus a loop-graph with linearly dependent loops is singular, analogous to a matrix with linearly dependent columns.

**Theorem 9** A loop-graph with $\sum_R c_R > 1$ is singular.

**Proof:** The graph of singletons has $\sum_R c_R = V$. To reduce the loop-graph to this, you need to split with empty separator at least $V - 1$ times. Only the outer regions can split, and each such split increases the total counting number by 1. (Welling et al., 2005). Hence it is impossible to get exactly $V$ starting from $\sum_R c_R > 1$ (Note that this proof is valid for any RG but not necessarily any SRG). **Another proof:** By Th.7 we must have $L > E - V + 1$. Additionally, it is known that the dimension of the cycle space is $E - V + 1$ (Diestel, 2000), so there must be a set of linearly dependent loops, analogously to a matrix with more columns than rows. □

Consequently, any loop-graph which is non-singular and has $\sum_R c_R = 1$ is *maximal* in the sense that any additional loop would make it singular. This type of loop-graph captures as many interactions as possible through the inclusion of loops, without adding spurious fixed points in the uniform case. For example, any loop-graph derived from an acyclic SRG is maximal. Additionally, any loop-graph corresponding to the inner faces of a planar graph is maximal, because any subset of loops has edges which are not shared, and $\sum_R c_R = 1$ by Euler's characteristic ($L - E + V = 1$).

| Error | Bethe | 1-star (+1) | 2-star (+1) | 3-star (+1) |
|---|---|---|---|---|
| Complete | .090 | .034 (.043) | .010 (.205) | .003 (.267) |
| Bipartite | .035 | .021 (.068) | .014 (.034) | .006 (.020) |

Table 1: Average estimation error of different region graphs over 20 random sets of edge potentials, for 6-node complete graphs and 20-node complete bipartite graphs. (+1) is with one extra region.

## 9 EXAMPLES

This section demonstrates these design principles on some standard problems. Consider a complete graph of pairwise factors. A region graph with all triples of nodes would be singular. Instead we should choose a subset of size $E - V + 1$. One choice is the EP-graph with a star base region (Minka & Qi, 2004). Letting node 1 be the root of the star, the cliques of the base region are $(1, i)$ where $i$ ranges over all nodes $> 1$. The outer regions add a single edge $(i, j)$, creating a triple $(1, i, j)$. After reduction, we get a loop-graph with these triples as the loop regions. By construction, this RG is non-singular and maximal ($\sum_R c_R = 1$). It is also closed under intersection (the intersection of any two outer regions corresponds to a clique in the base region), which means it is *totally connected* and *totally balanced* (Pakzad & Anantharam, 2003).

This construction can be generalized to higher width, using a width-$w$ star. Letting nodes $(1, ..., w)$ be the root of the star, the cliques of the base region are $(1, ..., w, i)$ where $i$ ranges over all nodes $> w$. The outer regions add a single edge $(i, j)$, creating a clique $(1, ..., w, i, j)$. After reduction, these cliques will be the outer regions. This RG is also non-singular, has $\sum_R c_R = 1$, and closed under intersection.

Fixing the order of nodes, we evaluated these region graphs on a 6-binary-node complete graph with random edge potentials (generated identically to Minka & Qi, 2004). In each trial we ran GBP with enough damping to ensure convergence, and measured the maximum error in the single-node marginals. The average performance over 20 trials is shown in table 1. The width-1-star has 1/3 the error of Bethe (at three times the computational cost), and each increment of width reduces the error by a further 1/3 (at 1% higher cost). To show that non-singularity is important, we added one arbitrary additional outer region $(2, ..., w+3)$ and ran this singular region graph on the same set of input graphs. This version requires much more damping, and the error is significantly worse (table 1). For $w = 1$, this extra region makes $\sum_R c_R = 2$, while for $w > 1$ we still have $\sum_R c_R = 1$ (perfect correlation despite singularity).

The star construction also works for non-complete graphs—just use the outer regions that correspond to factors. For example, consider complete bipartite graphs, such as figure 5a. Figure 5d corresponds exactly to a star RG with its root at the center node. Note that some of the edge regions in this construction will link variables that did not have factors (often essential for a good RG). Fixing the order of variables, we tested this approach on a 20-node graph (10 nodes in each partition) with random edge potentials as above. The average error over 20 trials is shown in table 1. The 1-star nearly halves the error of Bethe (at three times the cost) and each increment of width halves the error further (at double the cost).

Now consider grids of pairwise factors. By a similar construction to figure 1 (top right), there is an EP-graph which reduces to $n \times m$ boxes, overlapping at their sides. This RG is non-singular, has $\sum_R c_R = 1$, and closed under intersection. Message passing can be implemented efficiently by exploiting the internal structure of each box, which is a grid of factors plus a clique on all four sides, to represent incoming messages. For $3 \times 3$ boxes, this structure has treewidth 3, for $4 \times 3$ boxes the treewidth is 4, and for $4 \times 4$ boxes the treewidth is 5. By triangulating and splitting each box region, we obtain a region graph containing only small cliques, on which a standard GBP algorithm can run. Similar to the above experiments, there is a consistent increase in accuracy as you increase the treewidth of the boxes.

## 10 EXPERIMENTS

We have investigated the practical usefulness of non-singularity as a guiding principle in choosing region graphs. In particular we looked at using it in the context of region pursuit (Welling, 2004). We compared the original region pursuit, which is an iterative algorithm which adds regions one at a time on top of the Bethe approximation, to one in which regions are added only if the resulting RG is non-singular. We tested both versions on 7-binary-node complete pairwise Markov networks, where weights are sampled i.i.d. uniformly in the range $[0, .1]$, while biases are sampled uniformly in range $[0, .3]$ (weights have to be be small to ensure convergence). Regions consisting of triangles are considered, and the triangle which when added changes the free energy the most is chosen at each iteration (in fact we use an approximation to this, see Welling (2004) for details).

Figure 6 left panel shows the mean errors (over 50 trials) in the node marginals as we add triangles to the RG. The "Best" line uses the best triangle according to the "change in free energy" heuristic, while the "Worst" line uses the worst triangle. There are 35 tri-angles available. The region pursuit heuristic chooses good triangles in the beginning, but at around 15 triangles, adding more actually hurts the approximation. The "non-singular" line shows the alternative where triangles are not added if the resulting RG is singular, and figure 6 (right) shows the average number of triangles added (a maximum of 15 since the number of independent cycles is $E - V + 1 = 15$). The non-singular constraint stops the degradation. (Non-singularity was tested via the reduction process in the Th.8 proof.) Interestingly, at one point the "Best" line does better than the maximal non-singular RG. But lacking a way to detect this optimal stopping point, the non-singular constraint is the next best thing.

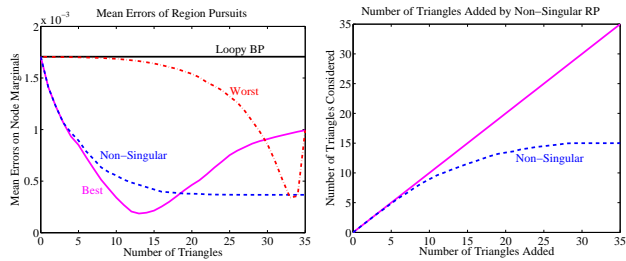

Figure 6: Modifying region pursuit by constraining region graphs to be non-singular.

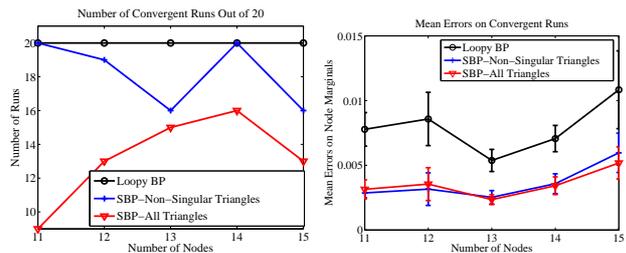

Figure 7: Convergence Properties of non-singular region graphs.

We have also found that non-singular RGs provide faster and more reliable convergence. We illustrate this in an experiment involving binary random networks between 11 and 15 nodes. Each edge appears independently with probability 0.75, and has weights i.i.d. from a Normal with zero mean and std. $1/\sqrt{n-1}$, while biases are zero mean, 1 std. Normal. We considered region graphs consisting of triangles, edges and nodes. In the control the triangles are chosen via region pursuit, in the alternative the triangles are considered in the same order, but not added if the resulting RG is singular. Both stopped when the RG reached 30 triangles. Figure 7 shows the result over 20 repetitions. Similar error rates were attained by both region graphs in this experiment.

## 11 DISCUSSION

While this paper has focused on discrete random variables, many of the results can be extended to continuous variables in general exponential families. This provides a generalization of EP to the Kikuchi case (see also Heskes & Zoeter, 2003).

Our criteria for good region graphs only involve structural information of the graph. However, the interaction strengths are also important in choosing good approximations. Criteria which incorporate this information seem a promising direction for progress.

The SRG formalism has already deepened the understanding of both EP and GBP algorithms, but many questions remain, for example: are there good region graphs which are not equivalent to EP-graphs? Can the notion of maximality for loop-graphs be generalized to other classes of region graphs? Are there additional design principles we can use, e.g. to choose among EP-graphs?